\documentclass[letterpaper, 10 pt, conference]{ieeeconf}  

\IEEEoverridecommandlockouts                              

\usepackage{graphics} 
\usepackage{epsfig} 
\usepackage{times} 
\usepackage{amsmath} 
\usepackage{amssymb}  
\usepackage{arydshln}
\usepackage{multirow}

\usepackage{xcolor}
\usepackage{soul}
\usepackage{fixltx2e}

\usepackage[nobiblatex]{xurl}

\usepackage{color,soul}

\usepackage{url}

\title{\LARGE \bf
Ablation Study on Features in Learning-based Joints Calibration of Cable-driven Surgical Robots
}

\author{Haonan Peng, 
Andrew Lewis,
Blake Hannaford \\[1ex]
\thanks{Haonan Peng, Andrew Lewis, and Blake Hannaford are with University of Washington, Seattle, WA 98195, USA.
        {\tt\small {penghn}@uw.edu}}%
}

\begin{document}
\bstctlcite{IEEEexample:BSTcontrol}

\maketitle
\thispagestyle{empty}
\pagestyle{empty}

\begin{abstract}
With worldwide implementation, millions of surgeries are assisted by surgical robots. The cable-drive mechanism on many surgical robots allows flexible, light, and compact arms and tools. However, the slack and stretch of the cables and the backlash of the gears introduce inevitable error from motor poses to joint poses, and thus forwarded to the pose and orientation of the end-effector. In this paper, a learning-based calibration using a deep neural network (DNN) is proposed, which reduces the unloaded pose RMSE of joints 1, 2, 3 to 0.3003$^{\circ}$, 0.2888$^{\circ}$, 0.1565 mm, and loaded pose RMSE of joints 1, 2, 3 to 0.4456$^{\circ}$, 0.3052$^{\circ}$, 0.1900 mm, respectively. Then, removal ablation and inaccurate ablation are performed to study which features of the DNN model contribute to the calibration accuracy. The results suggest that raw joint poses and motor torques are the most important features. For joint poses, the removal ablation shows that DNN model can derive this information from end-effector pose and orientation. For motor torques, the direction is much more important than amplitude. 
\end{abstract}

\section{Introduction}
Robot-assisted Minimally Invasive Surgery (RAMIS) brings the possibility to accurate and reliable operations between surgeons and surgical robots \cite{palep2009robotic}, with improved outcomes for patients \cite{sayari2019review}, \cite{peters2018review}. In abdominal RAMIS, the accuracy of millimeter level is preferred \cite{camarillo2004robotic}. In medical robotics mechanisms, the highly elongated form factors of endoscopic surgical instruments, and the necessary re-location of sensing and actuation to more proximal locations, impose fundamental limitations on mechanical accuracy. Cable drive mechanisms, traditionally used in robotic laparoscopic instruments, such as RAVEN-II \cite{hannaford2012raven} and da Vince Research Kit (dVRK) \cite{kazanzides2014open}, are problematic due to long runs of highly loaded drive cables, and their routing around numerous pulleys whose diameter is substantially below standard cable design guidelines. Recent efforts use machine learning calibration to reduce the error caused by the cable-driven mechanism and achieved competitive performance \cite{seita2018fast}, \cite{peng2020real}. However, what types of information features of the robots contribute to the learning calibration, as well as how the features contribute, remains undiscovered.

This paper studies use of machine learning models to correct for mechanical imperfections in the robot mechanism, specifical cable-driven transmissions from motors to joints, which reduce accuracy of control. In this work, we apply a deep neural network (DNN) to model these phenomena using data available in the control system and demonstrate its ability to improve accuracy of robot states in real-time with modest training data requirements. Furthermore, an ablation study on features of the DNN reveals insights into which types of real-time control data contribute to this performance improvement and which do not. The results imply that performance can be improved beyond the current limits of materials and mechanisms in demanding applications such as robotic manipulation in endoscopic (minimally invasive) surgery.   

\section{Related Work}
Model-based and learning-based methods were utilized to improve the cable-driven compromised accuracy of surgical robots. Model-based methods worked in an explainable manner, and benefitted from human knowledge of the kinematics and dynamics of the robot. Haghighipanah et al. \cite{haghighipanah2015improving} utilized Unscented Kalman Filter (UKF) to improve the estimation of positional joints of RAVEN-II. The UKF was based on the model of serial manipulator with elastic transmission and motor dynamics. The result showed considerable improvement in joints 2 and 3. Joint 1, however, saw a minor reduction in error due to stiff transmission. The performance of model-based method relies on accurate modeling of the system. Miyasaka et al. \cite{miyasaka2016hysteresis} developed a Bouc-Wen hysteresis model with Duhem operator on longitudinally loaded cable, the model had 9 parameters which should be determined by experiments. Due to joint coupling, it can be harder to accurately model the entire cable-driven robot.

In contrast, learning-based methods were favored in recent years. With sufficient data, proper model structure and training, learning-based methods bypassed the modeling of robot kinematics and dynamics, and achieved competitive performance. Hwang et al. \cite{hwang2020efficiently} proposed a recurrent neural network based calibration to reduce the positional error of dVRK's end-effector to sub-millimeter level, and super-human performance on auto peg transfer was achieved after calibration \cite{hwang2022automating}. Mahler et al. \cite{mahler2014learning} utilized Gaussian Process Regression to improve the accuracy of the end-effector of RAVEN-II. In the experiments, it was also found that including velocity as an input feature helped further reduce calibration error significantly, which posed a question on which kinds of features could be helpful in learing-based calibration of cable-driven surgical robots. 

By analogy with ablative brain surgery \cite{franzini2019ablative}, ablation study in Artificial Intelligence removes certain components of the AI model and compares the performance, to evaluate and understand the contribution of the components \cite{cohen1989toward}, \cite{newell1975tutorial}. A unit ablation study on computer vision classification accomplished by Meyes et al. \cite{meyes2019ablation} suggested that the more weight distribution changed during training, the more contributive that unit was. Kauchak et al. \cite{kauchak2014text} proposed an ablation study on features in a learning-based medical text difficulty classification. When removed, some features caused more changes in performance than other features.

In our paper, on a robotics regression task, we found that removal ablation - removing features in both training and testing, was not sufficient to determine whether a feature was informative. Thus we introduced inaccurate ablation - adding noise to features only in testing, and compared the results. We also tried to use human knowledge to explain the different results of the two ablation methods, and further explored the manner in how the features contribute to the task.

\section{Methods} 

\subsection{Features in the Robot State} 

\begin{figure*}
\centering
\vspace{1.5em}
\includegraphics[width=0.85\textwidth]{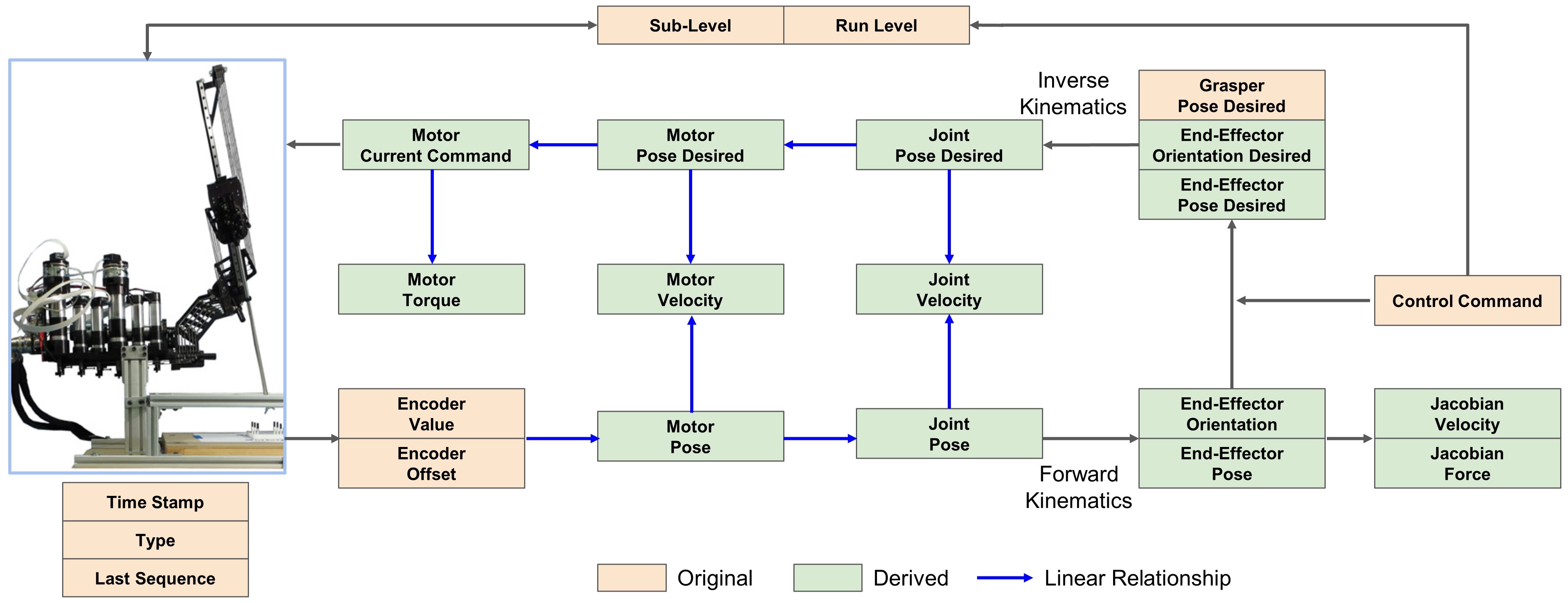}
\vspace{-1.em}
\caption{Relationship in the robot states.} 
\vspace{-1.5em}
\label{Fig_ravenstate}
\end{figure*}

To establish a learning-based calibration with competitive performance, comprehensive input features should be provided to the machine learning model. For the software system of RAVEN-II, a ROS message - 'ravenstate', integrates the measurements, control command, and state estimation of the robot. Details of 'ravenstate' are shown in Table \ref{Tab_ravenstate}. Robot states messages are published at 1000Hz and are down-sampled to ~25Hz as the input of the following DNN model.

The features in the robot states are not all independent and original, as shown in Fig. \ref{Fig_ravenstate}. For example, motor poses are derived from encoder values and offsets, with a linear relationship. On the other hand, end-effector pose and orientation are derived from joint poses by forward kinematics, which is a highly non-linear relationship. Recall that due to the cable-driven mechanism, all encoders of RAVEN-II robot are placed on the motors at the backend and there is no joint encoder. The joint poses are derived from motor poses and thus contain a considerable amount of error because of the slack and stretch of the cables. This error is also forwarded to end-effector pose and orientation by forward kinematics, which significantly reduces the open-loop accuracy of the robot.

\begin{table*}[h]
\vspace{1.em}
\caption{\vspace{-0.3em} Features in RAVEN-II State}
\vspace{-1.5em}
\begin{center}
\begin{tabular}{c|c|p{0.6\linewidth}}
\hline
\textbf{Features}                       & \textbf{Dimension} & \multicolumn{1}{c}{\textbf{Description}}                                                                                                             \\ \hline
Time Stamp                       & 1         & Time stamp of the ROS message, converted to second.                                                                                         \\
Run Level                        & 1         & An integer indicates major run level of the robot, including homing,  pedal up (pausing), pedal down (operating), and E-stop.      \\
Sub-level                        & 1         & An integer indicates subordinate run levels under major run levels.                                                                                         \\
Last Sequence                    & 1         & Sequence number of teleoperation commands.                                                                                                                              \\
Arm Type                             & 1         & An integer indicates left/right arm.                                                                                                                               \\ \hline
End-Effector Pose                & 3         & (x, y, z) cartesian pose of the end-effector, derived from joint poses using forward kinematics.                                            \\
End-Effector Pose Desired        & 3         & Desired (x, y, z) cartesian pose of the end-effector, derived from control command.                                                         \\
End-Effector Orientation         & 9         & 3 $\times$ 3 rotation matrix of the orientation of the end-effector, derived from joint poses using forward kinematics. \\
End-Effector Orientation Desired & 9         & 3 $\times$ 3 rotation matrix of the desired orientation of the end-effector, derived from control command.              \\ \hline
Encoder Value                    & 7         & Encoder values of the motors of the 7 joints.                                                                                               \\
Encoder Offset                   & 7         & Encoder offsets of the motors of the 7 joints, calibrated during homing, and stay constant.                                \\ \hline
Motor Pose                       & 7         & Motor poses of the 7 joints, derived from encoder values and offsets.                                                                       \\
Motor Pose Desired               & 7         & Desired motor pose of the 7 joints, derived from desired joint poses.                                                                       \\
Motor Velocity                   & 7         & Motor velocity of the 7 joints, derived from motor poses and desired motor poses.                                                           \\ \hline
Joint Pose                       & 7         & Angle/translation of the 7 joints, derived from motor poses.                                                                              \\
Joint Pose Desired               & 7         & Desired joint pose of the 7 joints, derived from joint control command, or Cartesian control command and inverse kinematics.                \\
Joint Velocity                   & 7         & Joint velocity of the 7 joints, derived from joint poses and desired joint poses.                                                           \\ \hline
Motor Current Command                 & 7         & DC current commands of the motors, derived from current and desired motor poses.                                                                            \\
Motor Torque                     & 7         & Torques of the motors, derived from the measured motor currents.                                                                            \\ \hline
Jacobian Velocity                & 6         & Jacobian velocity of the end effector.                                                                         \\
Jacobian Force                   & 6         & Jacobian force of the end effector.                                                                             \\ \hline
Desired Grasper Pose             & 1         & A float number represents open/close command of the grasper.                                                                                                                              \\ \hline
\end{tabular}
\end{center}
\vspace{-2.5em}
\label{Tab_ravenstate}
\end{table*}

\subsection{Learning-based Calibration} \label{schar_DNN}
In order to use the machine learning method to provide more accurate joint poses than the original state estimation, a DNN model is setup. The DNN model takes the flattened 'ravenstate' ROS message as input. However, the output is the error of the first 3 joint poses, instead of the joint poses themselves, which gives better generalization ability and robustness. Learning calibrated joint poses can be obtained by adding the joint pose errors back to the robot's original inaccurate joint poses.

The structure and hyper-parameters are empirically chosen. Too deep layers with more hidden units tend to have compromised performance. Adam optimizer is used \cite{kingma2014adam} \cite{reddi2019convergence}, with exponential decay rate for the 1st and 2nd moment estimates of 0.9 and 0.999, respectively. The learning rate is fixed at 0.0005 with 600 training epochs and batch size of 1024. The Sigmoid activation function is used in all layers except for the output layer which uses linear activation. Mean squared error is used as the loss function. L1 and L2 regularization is applied on each hidden layer kernel with factors of 1e-5 and 1e-4, respectively. L2 regularization is also applied on each hidden layer's bias and output with factors of 1e-4 and 1e-5, respectively.

\begin{figure}
\centering
\vspace{0.5em}
\includegraphics[width=0.36\textwidth]{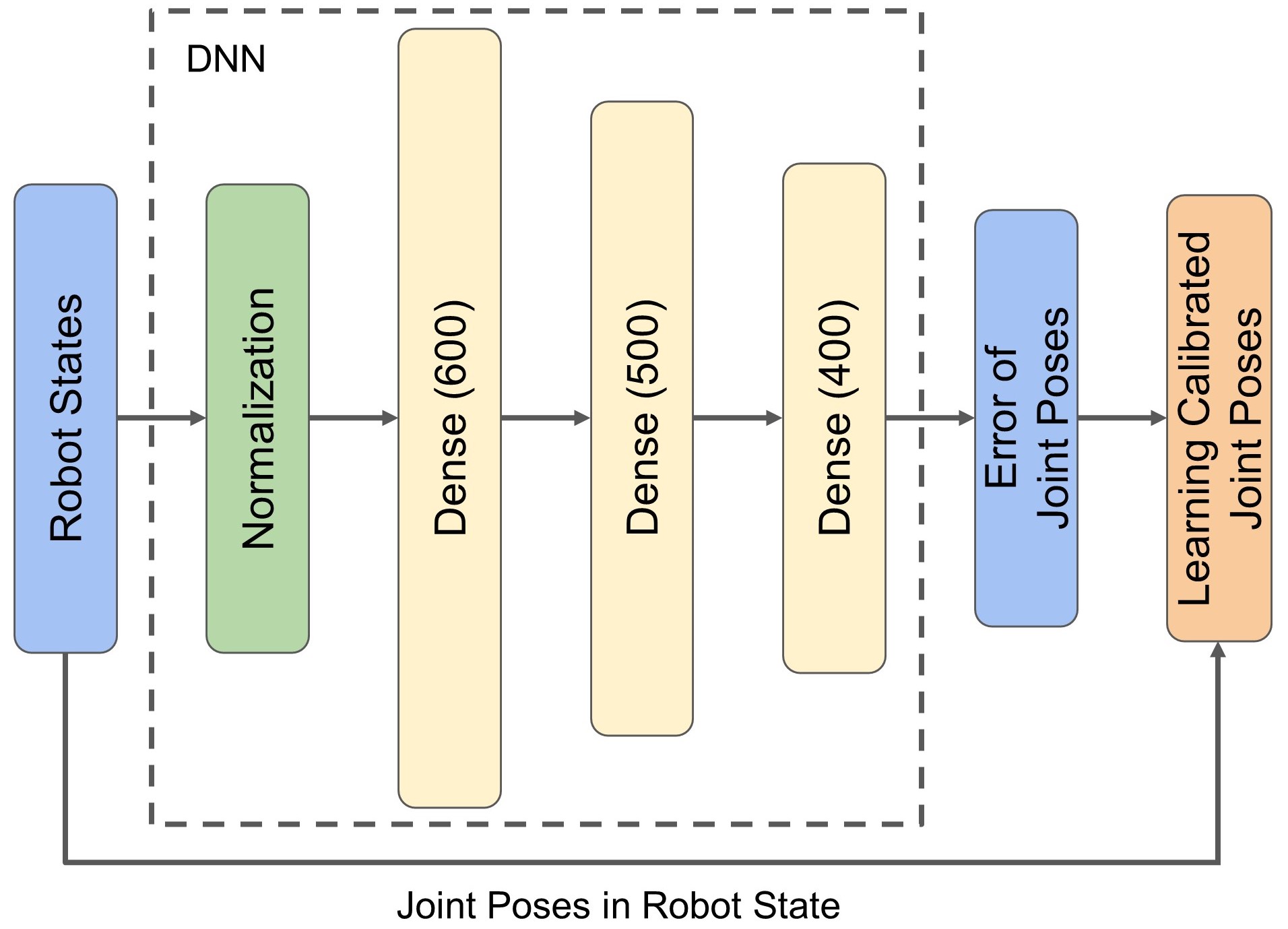}
\vspace{-1.em}
\caption{Model of the DNN in learning calibration. The output of the DNN is the error of the joint poses and is added back to the original joint pose estimation for correction.}
\vspace{-1.5em}
\label{DNN_model}
\end{figure}

\subsection{Trajectory of Training and Testing} \label{schar_traj}

\begin{figure*}
\centering
\vspace{1.5em}
\includegraphics[width=0.85\textwidth]{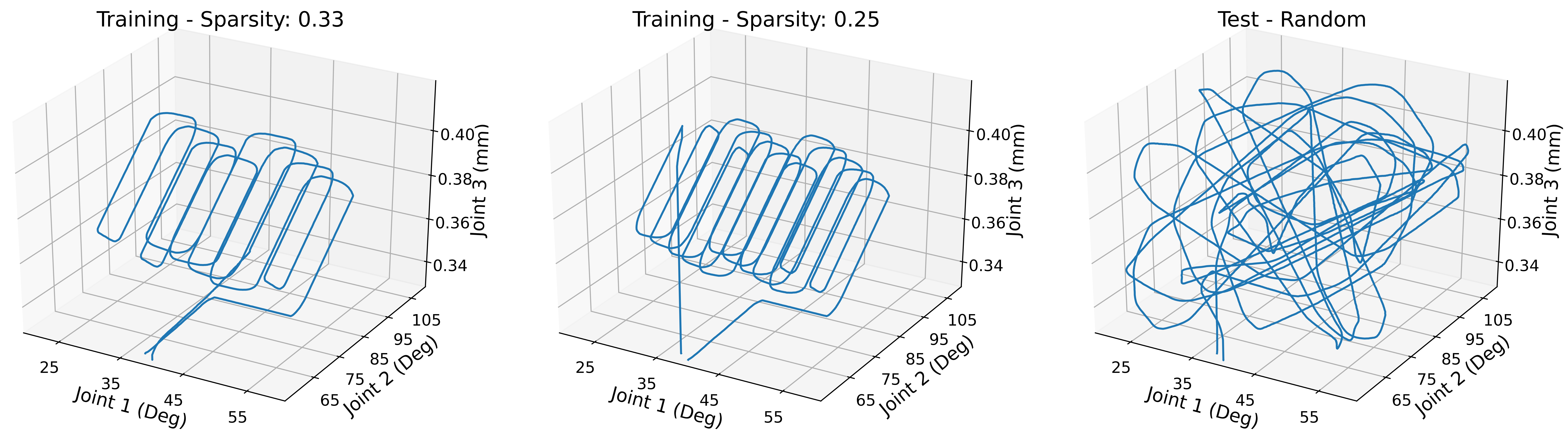}
\vspace{-1.em}
\caption{Examples of zig-zag training and random test trajectories.} 
\vspace{-1.5em}
\label{Fig_trajs}
\end{figure*}

With the DNN model in \label{schar_DNN}, data is needed to train the model. To ensure reliable ground truth joint pose measurement, the first 3 joints, positional joints, are studied in this paper. Zig-zag trajectories cover the workspace efficiently with desired sparsity and are thus used as training trajectories. All the trajectories in this paper are in joint space instead of Cartesian space.

To prove the generalization ability of the model trained with zig-zag trajectories, random sinusoidal trajectories are used for testing. In each cycle, the velocity and target pose of each joint are randomly chosen. 

\subsection{Ablation Study on Features} \label{schar_ablation_study}
To study how the features of the robot states contribute to the performance of learning calibration, different ablation studies are performed. In terms of the ablation method, removal ablation and inaccurate ablation are used. On the other hand, in terms of ablation target, feature type ablation and joint ablation are used. Thus, the method and target of ablation combine into 4 ablation studies: 1) removal ablation on feature types, 2) inaccurate ablation on feature types, 3) removal ablation on joints, and 4) inaccurate ablation on joints.

\textbf{Removal ablation} drops the target features both in training and testing, which results in a reduction of the input dimension of the learning model. If the target features have no or minor contribution to the learning calibration, the removal of these features should not undermine the performance considerably. If the target features do have a major contribution, the removal of these features will force the model to learn to derive the missing information from the remaining features. For example, end-effector poses and orientation can be derived from joint poses using forward kinematics, and reversely using inverse kinematics, though both relations are highly non-linear for a 7-joint robot.

\textbf{Inaccurate ablation} only adds noise to target features during testing, while these target features stay accurate during training. Compared with removal ablation, inaccurate ablation does not allow the learning model to know the malfunctioning feature during training, and thus the model is not forced to learn to derive the related information from other accurate features. If the target features have a major contribution, the learning model will tend to rely on these features during training. However, during testing, these features are inaccurate and should thus cause a decline in performance. \textbf{[add equation here]}

\textbf{Ablation on feature types} focuses on chosen types of features, such as motor poses and joint poses. Because there are 22 different types of features in the robot states and some features are linear-related, ablation on different feature types is performed in groups of features. Once a feature is chosen for ablation, that feature of each joint is removed or with noise added.

\textbf{Ablation on joints} focuses on features of the chosen joint. This can help study the coupling of the joints. For example, if removal ablation of joint 1 is performed, all the features regarding joint 1 are removed, including (desired) joint pose/velocity, (desired) motor pose/velocity/torque, and motor current command. Although end-effector pose \& orientation and Jacobian velocity \& force are also related to joint 1, they are not included due to high non-linear relation.

\section{Experiments and Results}

\subsection{Experiments Setup} \label{schar_exp_setup}
A RAVEN-II surgical robot was used to perform the experiments. Only the left arm was used and optical incremental joint encoders were placed on the first 3 joints to provide accurate measurement of joint poses. Joint 1 and joint 2 are rotational, and thus Avago Technologies AEDA-3300 encoders were used, with the resolution of 80000 counts after quadrature decode. Joint 3 is translational, and thus Mercury II 1600 was used, with the resolution of 5 {\textmu}m. The external joint encoders were calibrated to joint poses during RAVEN's homing procedure, in which RAVEN explores the mechanical hard limit of each joint.

The first three Raven arm joints have substantially different mechanical properties. First, like any serial arm, the loads (including both gravity torques and other loads) on each joint decrease from 1 to 3. Second, the cable lengths in joints 1 to 3 are substantially different. Joint 1 has a single alpha-wrapped pulley joint and essentially zero free cable length. In contrast, joints 2 and 3 have increasing cable lengths.

Joint control commands were sent through Collaborative Robotics Toolkit (CRTK) API \cite{su2020collaborative}, and the robot states were obtained from the ROS topic 'ravenstate'.

\begin{figure}
\centering
\vspace{0.5em}
\includegraphics[width=0.4\textwidth]{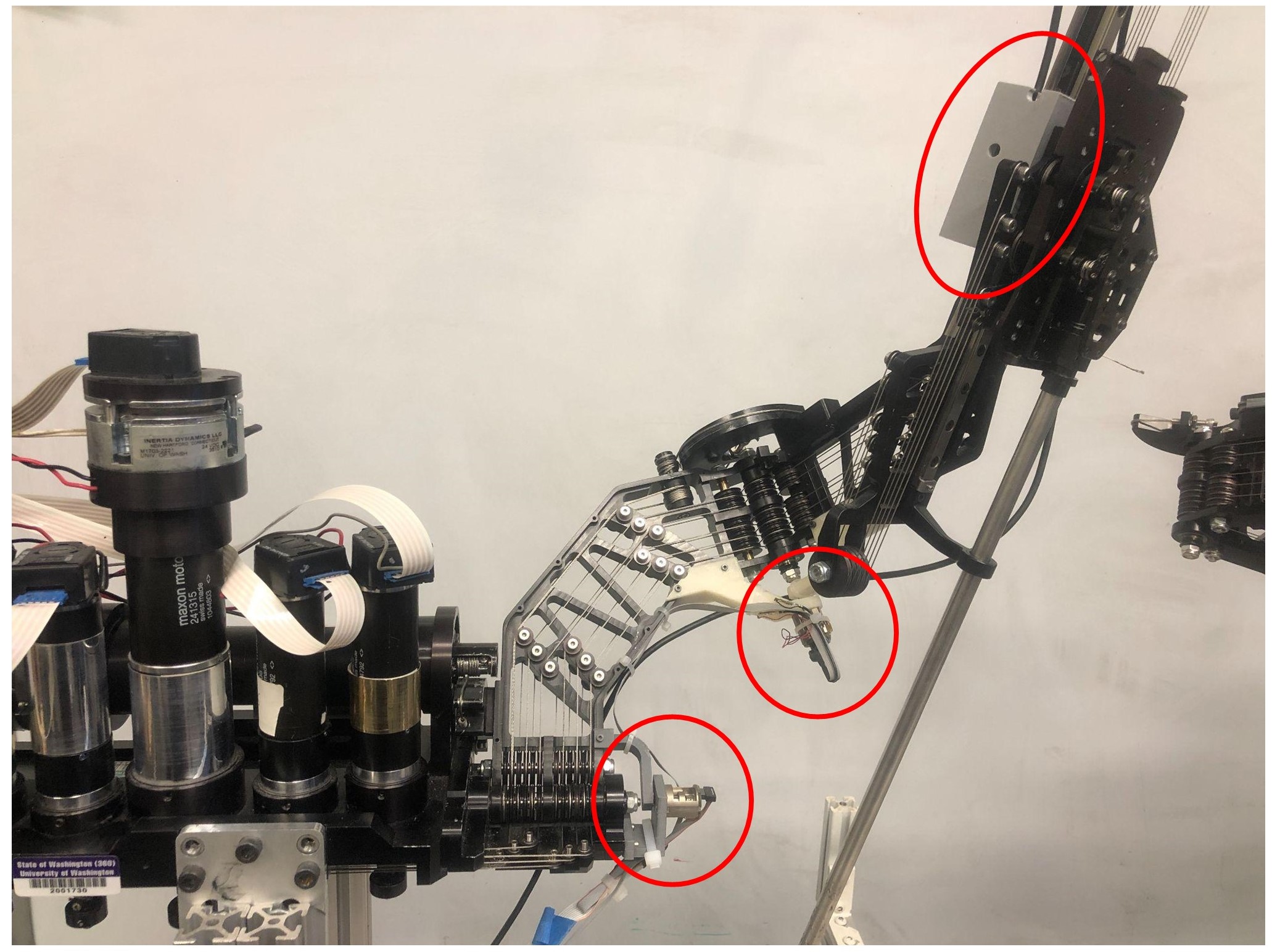}
\vspace{-1.em}
\caption{External joint encoders are mounted on the first 3 joints of RAVEN. Actual joint poses are measured to provide ground truth during training and testing of the learning model.} 
\vspace{-1.5em}
\label{exp_setup}
\end{figure}

\subsection{Datasets} \label{sch_datasets}
As mentioned in \ref{schar_traj}, 5 zig-zag trajectories with different sparsity of 1/2, 1/3, 1/4, 1/5, and 1/6, were used as the training set. The 5 trajectories were combined when training and it took 3244 seconds to collect the data. The joint limits were set as [20, 60] degrees, [55, 110] degrees, and [0.33, 0.42] meters for joints 1, 2, and 3, respectively. Overall, training set consisted of around 68000 data pairs. 

To study the long-term accuracy of the learning calibration, the test sets were collected by continuously running random sinusoidal trajectories for 6 hours, which resulted in around 360000 data pairs. The training and test sets were collected two times, one without and load and one with a 500g load applied on the end-effector.

\subsection{Ablation Study: Removal of Features} 

\begin{table*}[h]
\vspace{1.em}
\caption{\vspace{-0.3em} Learning Calibration Error with Removal of Features}
\vspace{-2.5em}
\begin{center}
\begin{tabular}{c||rrrrrr||rrrrrr}
\hline
                               & \multicolumn{6}{c||}{No Load}                                                                                                                                                                               & \multicolumn{6}{c}{500g Load}                                                                                                                                                                             \\ \cline{2-13} 
Removed Features               & \multicolumn{2}{c|}{Joint 1 (Deg)}                                       & \multicolumn{2}{c|}{Joint 2 (Deg)}                                       & \multicolumn{2}{c||}{Joint 3 (mm)}                    & \multicolumn{2}{c|}{Joint 1 (Deg)}                                       & \multicolumn{2}{c|}{Joint 2 (Deg)}                                       & \multicolumn{2}{c}{Joint 3 (mm)}                    \\
                               & \multicolumn{1}{c}{RMSE} & \multicolumn{1}{c|}{Peak}                     & \multicolumn{1}{c}{RMSE} & \multicolumn{1}{c|}{Peak}                     & \multicolumn{1}{c}{RMSE} & \multicolumn{1}{c||}{Peak} & \multicolumn{1}{c}{RMSE} & \multicolumn{1}{c|}{Peak}                     & \multicolumn{1}{c}{RMSE} & \multicolumn{1}{c|}{Peak}                     & \multicolumn{1}{c}{RMSE} & \multicolumn{1}{c}{Peak} \\ \hline
Before Calibration             & 2.1150                   & \multicolumn{1}{r|}{3.2501}                   & 7.9019                   & \multicolumn{1}{r|}{10.8255}                  & 11.6895                  & 12.9004                   & 2.1787                   & \multicolumn{1}{r|}{3.3802}                   & 7.9144                   & \multicolumn{1}{r|}{10.8637}                  & 11.2281                  & 12.3236                  \\
Bias Removal                   & 0.9324                   & \multicolumn{1}{r|}{1.7420}                   & 1.0990                   & \multicolumn{1}{r|}{3.1237}                   & 0.5266                   & 1.2192                    & 1.0158                   & \multicolumn{1}{r|}{1.9028}                   & 1.1147                   & \multicolumn{1}{r|}{3.1945}                   & 0.5034                   & 1.1180                   \\
\textit{\textbf{All Features}} & \textit{\textbf{0.3003}} & \multicolumn{1}{r|}{\textit{\textbf{0.5902}}} & \textit{\textbf{0.2888}} & \multicolumn{1}{r|}{\textit{\textbf{1.2669}}} & \textit{\textbf{0.1565}} & \textit{\textbf{0.6409}}  & \textit{\textbf{0.4456}} & \multicolumn{1}{r|}{\textit{\textbf{0.7721}}} & \textit{\textbf{0.3052}} & \multicolumn{1}{r|}{\textit{\textbf{1.4299}}} & \textit{\textbf{0.1900}} & \textit{\textbf{0.7513}} \\ \hline
Operating Status               & 0.3278                   & \multicolumn{1}{r|}{0.6119}                   & 0.3331                   & \multicolumn{1}{r|}{1.4652}                   & 0.1556                   & 0.5693                    & 0.4602                   & \multicolumn{1}{r|}{0.7792}                   & 0.3709                   & \multicolumn{1}{r|}{1.5928}                   & 0.1765                   & 0.7587                   \\
Joint Poses           & 0.3222                   & \multicolumn{1}{r|}{0.6414}                   & 0.3050                   & \multicolumn{1}{r|}{1.4229}                   & 0.1268                   & 0.5153                    & 0.3781                   & \multicolumn{1}{r|}{0.7750}                   & 0.3026                   & \multicolumn{1}{r|}{1.4209}                   & 0.1700                   & 0.6823                   \\
End-effector           & 0.3236                   & \multicolumn{1}{r|}{0.5842}                   & 0.2833                   & \multicolumn{1}{r|}{1.2922}                   & 0.1538                   & 0.6330                    & 0.4587                   & \multicolumn{1}{r|}{0.7464}                   & 0.2860                   & \multicolumn{1}{r|}{1.3915}                   & 0.1883                   & 0.7539                   \\
\textbf{All Poses}             & \textbf{0.8645}          & \multicolumn{1}{r|}{\textbf{3.8587}}          & \textbf{1.2448}          & \multicolumn{1}{r|}{\textbf{3.8877}}          & \textbf{0.2539}          & \textbf{1.1198}           & \textbf{1.1766}          & \multicolumn{1}{r|}{\textbf{5.0998}}          & \textbf{1.8213}          & \multicolumn{1}{r|}{\textbf{5.6556}}          & \textbf{0.2852}          & \textbf{1.1605}          \\
\textbf{Torque}        & 0.2968                   & \multicolumn{1}{r|}{0.5833}                   & \textbf{0.4877}          & \multicolumn{1}{r|}{\textbf{1.9435}}          & \textbf{0.3451}          & \textbf{1.2021}           & 0.4196                   & \multicolumn{1}{r|}{0.7551}                   & \textbf{0.5170}          & \multicolumn{1}{r|}{\textbf{2.0554}}          & \textbf{0.3533}          & \textbf{1.2582}          \\
\textbf{Velocity}      & 0.3218                   & \multicolumn{1}{r|}{0.6028}                   & 0.3456          & \multicolumn{1}{r|}{1.4280}          & 0.1642                   & 0.6710                    & 0.4566                   & \multicolumn{1}{r|}{0.7682}                   & 0.3840          & \multicolumn{1}{r|}{1.6096}          & 0.2085                   & 0.8035                   \\
Jacobian               & 0.3172                   & \multicolumn{1}{r|}{0.5977}                   & 0.2944                   & \multicolumn{1}{r|}{1.3160}                   & 0.1529                   & 0.6272                    & 0.4478                   & \multicolumn{1}{r|}{0.7680}                   & 0.3294                   & \multicolumn{1}{r|}{1.5004}                   & 0.1944                   & 0.7548                   \\ \hline
Joint 1                        & 0.3439                   & \multicolumn{1}{r|}{\textbf{0.7868}}                   & 0.3019                   & \multicolumn{1}{r|}{1.3246}                   & 0.1470                   & 0.6352                    & 0.5195                   & \multicolumn{1}{r|}{\textbf{1.0623}}                   & 0.3320                   & \multicolumn{1}{r|}{1.4378}                   & 0.1718                   & 0.7415                   \\
\textbf{Joint 2}               & 0.3548                   & \multicolumn{1}{r|}{0.6628}          & \textbf{1.3047}          & \multicolumn{1}{r|}{\textbf{3.9894}}          & \textbf{0.2833}          & \textbf{0.9104}           & 0.5323                   & \multicolumn{1}{r|}{0.8727}          & \textbf{1.4744}          & \multicolumn{1}{r|}{\textbf{4.2773}}          & \textbf{0.3752}          & \textbf{1.2024}          \\
Joint 3                        & 0.3307                   & \multicolumn{1}{r|}{0.5934}                   & 0.3461                   & \multicolumn{1}{r|}{1.4426}                   & 0.1684                   & 0.6671                    & 0.4806                   & \multicolumn{1}{r|}{0.7712}                   & 0.3476                   & \multicolumn{1}{r|}{1.5084}                   & 0.2033                   & 0.7712                   \\ \hline

\multicolumn{13}{l}{* i) Errors with an increase larger than 30\% are marked bold; ii) Bias removal used the training sets to compute the average difference between}\\
\multicolumn{13}{l}{$~~$joint poses in robot states and ground truth, and then added compensation constants to joint poses in the test set, then errors were evaluated}\\
\multicolumn{13}{l}{$~~$using ground truth in the test set.}\\

\end{tabular}
\end{center}
\vspace{-2.5em}
\label{tab_removal_ablation}
\end{table*}

\begin{table*}[h]
\vspace{1.em}
\caption{\vspace{-0.3em} Learning Calibration Error with Inaccurate Features}
\vspace{-2.5em}
\begin{center}
\begin{tabular}{c||rrrrrr||rrrrrr}
\hline
                               & \multicolumn{6}{c||}{No Load}                                                                                                                                                                               & \multicolumn{6}{c}{500g Load}                                                                                                                                                                             \\ \cline{2-13} 
Inaccurate Features            & \multicolumn{2}{c|}{Joint 1 (Deg)}                                       & \multicolumn{2}{c|}{Joint 2 (Deg)}                                       & \multicolumn{2}{c||}{Joint 3 (mm)}                    & \multicolumn{2}{c|}{Joint 1 (Deg)}                                       & \multicolumn{2}{c|}{Joint 2 (Deg)}                                       & \multicolumn{2}{c}{Joint 3 (mm)}                    \\
                               & \multicolumn{1}{c}{RMSE} & \multicolumn{1}{c|}{Peak}                     & \multicolumn{1}{c}{RMSE} & \multicolumn{1}{c|}{Peak}                     & \multicolumn{1}{c}{RMSE} & \multicolumn{1}{c||}{Peak} & \multicolumn{1}{c}{RMSE} & \multicolumn{1}{c|}{Peak}                     & \multicolumn{1}{c}{RMSE} & \multicolumn{1}{c|}{Peak}                     & \multicolumn{1}{c}{RMSE} & \multicolumn{1}{c}{Peak} \\ \hline
Before Calibration             & 2.1150                   & \multicolumn{1}{r|}{3.2501}                   & 7.9019                   & \multicolumn{1}{r|}{10.8255}                  & 11.6895                  & 12.9004                   & 2.1787                   & \multicolumn{1}{r|}{3.3802}                   & 7.9144                   & \multicolumn{1}{r|}{10.8637}                  & 11.2281                  & 12.3236                  \\
Bias Removal                   & 0.9324                   & \multicolumn{1}{r|}{1.7420}                   & 1.0990                   & \multicolumn{1}{r|}{3.1237}                   & 0.5266                   & 1.2192                    & 1.0158                   & \multicolumn{1}{r|}{1.9028}                   & 1.1147                   & \multicolumn{1}{r|}{3.1945}                   & 0.5034                   & 1.1180                   \\
\textit{\textbf{No Inaccuracy}} & \textit{\textbf{0.3003}} & \multicolumn{1}{r|}{\textit{\textbf{0.5902}}} & \textit{\textbf{0.2888}} & \multicolumn{1}{r|}{\textit{\textbf{1.2669}}} & \textit{\textbf{0.1565}} & \textit{\textbf{0.6409}}  & \textit{\textbf{0.4456}} & \multicolumn{1}{r|}{\textit{\textbf{0.7721}}} & \textit{\textbf{0.3052}} & \multicolumn{1}{r|}{\textit{\textbf{1.4299}}} & \textit{\textbf{0.1900}} & \textit{\textbf{0.7513}} \\ \hline
Operating Status               & 0.3173                   & \multicolumn{1}{r|}{0.6023}                   & 0.3288                   & \multicolumn{1}{r|}{1.4112}                   & 0.1533                   & 0.6283                    & 0.4632                   & \multicolumn{1}{r|}{0.7848}                   & 0.3777                   & \multicolumn{1}{r|}{1.5940}                   & 0.1873                   & 0.7327                   \\
\textbf{Joint Poses}   & \textbf{6.7101}          & \multicolumn{1}{r|}{\textbf{12.2552}}         & \textbf{8.4935}          & \multicolumn{1}{r|}{\textbf{15.9875}}         & 0.1566                   & 0.6949                    & \textbf{6.7082}          & \multicolumn{1}{r|}{\textbf{12.5252}}         & \textbf{8.7467}          & \multicolumn{1}{r|}{\textbf{16.6830}}         & 0.1919                   & 0.8043                   \\
\textbf{End-effector}  & 0.3273                   & \multicolumn{1}{r|}{\textbf{0.7820}}                   & 0.3515          & \multicolumn{1}{r|}{\textbf{1.7050}}          & 0.1559                   & 0.6716                    & 0.4716                   & \multicolumn{1}{r|}{0.9822}                   & \textbf{0.3967}          & \multicolumn{1}{r|}{\textbf{1.8873}}          & 0.1886                   & 0.7529                   \\
\textbf{All Poses}             & \textbf{6.7187}          & \multicolumn{1}{r|}{\textbf{12.3596}}         & \textbf{8.5073}          & \multicolumn{1}{r|}{\textbf{16.1375}}         & 0.1593                   & 0.7380                    & \textbf{6.7015}          & \multicolumn{1}{r|}{\textbf{12.5493}}         & \textbf{8.7489}          & \multicolumn{1}{r|}{\textbf{16.7808}}         & 0.1939                   & 0.8142                   \\
\textbf{Torque}        & 0.3163                   & \multicolumn{1}{r|}{0.6227}                   & 0.3425          & \multicolumn{1}{r|}{1.5971}          & 0.1635                   & 0.7677                    & 0.4656                   & \multicolumn{1}{r|}{0.8259}                   & 0.3901          & \multicolumn{1}{r|}{1.7438}          & 0.1952                   & 0.8456                   \\
Velocity             & 0.3143                   & \multicolumn{1}{r|}{0.6043}                   & 0.3299                   & \multicolumn{1}{r|}{1.4404}                   & 0.1534                   & 0.6328                    & 0.4620                   & \multicolumn{1}{r|}{0.7879}                   & 0.3787                   & \multicolumn{1}{r|}{1.6168}                   & 0.1873                   & 0.7386                   \\
Jacobian              & 0.3170                   & \multicolumn{1}{r|}{0.6040}                   & 0.3290                   & \multicolumn{1}{r|}{1.4140}                   & 0.1534                   & 0.6321                    & 0.4625                   & \multicolumn{1}{r|}{0.7912}                   & 0.3804                   & \multicolumn{1}{r|}{1.5996}                   & 0.1871                   & 0.7259                   \\ \hline
\textbf{Joint 1}               & \textbf{6.7116}          & \multicolumn{1}{r|}{\textbf{12.2601}}         & 0.3297                   & \multicolumn{1}{r|}{1.4271}                   & 0.1536                   & 0.6349                    & \textbf{6.7102}          & \multicolumn{1}{r|}{\textbf{12.5001}}         & 0.3798                   & \multicolumn{1}{r|}{1.6142}                   & 0.1875                   & 0.7520                   \\
\textbf{Joint 2}               & 0.3159                   & \multicolumn{1}{r|}{0.6085}                   & \textbf{8.5001}          & \multicolumn{1}{r|}{\textbf{16.0838}}         & 0.1564                   & 0.7047                    & 0.4623                   & \multicolumn{1}{r|}{0.7920}                   & \textbf{8.7420}          & \multicolumn{1}{r|}{\textbf{16.6404}}         & 0.1921                   & 0.8032                   \\
Joint 3                        & 0.3160                   & \multicolumn{1}{r|}{0.6054}                   & 0.3305                   & \multicolumn{1}{r|}{1.4392}                   & 0.1562                   & 0.6754                    & 0.4628                   & \multicolumn{1}{r|}{0.7890}                   & 0.3799                   & \multicolumn{1}{r|}{1.6141}                   & 0.1893                   & 0.7620                   \\ \hline

\multicolumn{13}{l}{* i) Errors with an increase larger than 30\% are marked bold.}\\

\end{tabular}
\end{center}
\vspace{-2.5em}
\label{tab_inaccurate_ablation}
\end{table*}

As mentioned in \ref{schar_ablation_study}, removal ablation was performed by removing the target feature in both training and testing. The joint pose errors after learning calibration are shown in Table \ref{tab_removal_ablation}. Each entry regarding learning calibration was an average of 5 repetitive training and testing with different global random seeds, which applies to all experiments in this paper. Standard deviations were generally smaller than 5\% and were thus omitted. Two baselines were used as reference: error before calibration and error after constant bias removal. Removal ablation study on types of features was accomplished in groups: 1) \textbf{operating status} - including time stamp, run level, sub-level, last sequence, arm type and desired grasper pose; 2) \textbf{motor \& joint poses} - includes motor pose \& desired, joint pose \& desired; 3) \textbf{end-effector pose and orientation} - includes end-effector pose \& desired, end-effector orientation \& desired; 4) \textbf{all poses} - including all features in 2) and 3); 5) \textbf{velocity} - includes motor velocity and joint velocity; 6) \textbf{torque} - includes motor current command and motor torque; 7) \textbf{Jacobian} - includes Jacobian velocity and Jacobian force. One group of features was removed at a time and the calibration error was compared by using all features to train the DNN model. Encoder values, offsets, and the 4th motor pose were excluded from all experiments unless stated otherwise (details in \ref{schar_homing_inconsist}). 

It can be seen that the learning-based method improved the accuracy of joint pose estimation significantly. But the performance differed when certain features were not provided. In terms of different feature types, the biggest error in joint 1 was seen when all poses related features were removed, compared with using all features, the RMSE increased for +0.564$^{\circ}$ without load and +0.731$^{\circ}$ with 500g load, while the removal of the rest type of feature did not cause a considerable increase in calibration error. For joint 2, the removal of all poses related features also introduced the largest increase in error, +0.956$^{\circ}$ without load and +1.516$^{\circ}$ when loaded. Removal of torque and velocity related features also caused a certain increase in error, but was much less than removal of poses related features. Unlike rotational joints 1 and 2, for translational joint 3, the largest error increase, +0.189 mm (unloaded) and +0.163 mm (loaded), was seen by the removal of torque related features. A smaller increase in error was also caused by the removal of velocity features.

When all features related to joint 1 were removed, no major error increase was observed in all 3 joints. However, when removing joint 2 features, not only did the joint 2 RMSE increase by +1.015$^{\circ}$ (unloaded) and +1.169$^{\circ}$ (loaded), joint 3 RMSE also increased by +0.127 mm and +0.185 mm without and with load, respectively. In contrast, removing joint 3 features did not increase the error on joint 3, as well as the other 2 joints.

\subsection{Ablation Study: Inaccuracy of Features}
The inaccuracy ablation study was constructed similarly to removal ablation, the difference was instead of removing the target features, large white noises with range of 2 times the standard deviation were added to these features during testing, but not during training. The result is shown in Table \ref{tab_inaccurate_ablation}. In terms of different feature types, inaccurate joint poses related features caused a tremendous error that was even larger than the error before the calibration on joints 1 and 2. However, the accuracy of joint 3 was almost unaffected by the noise added. Adding noise to both joint and end-effector poses related features caused similar error increases to only adding noise to joint pose related features. In contrast, only adding noise to end-effector features did not increase the error considerably, except for joint 2, but it was still much less than the error caused by adding noise to joint features. Compared to inaccurate pose features, inaccurate torque only increased the error of joint 2 slightly with +0.054$^{\circ}$ and 0.085$^{\circ}$ when unloaded and loaded, respectively.

When all features of joint 1 were inaccurate, the error only increased significantly on joint 1. Similarly, inaccurate joint 2 features also only undermined learning calibration on joint 2. However, in contrast, inaccurate joint 3 features had no obvious influence on the calibration accuracy, even on joint 3 itself.

\subsection{Effectiveness of Torques}
Comparing Table \ref{tab_removal_ablation} and \ref{tab_inaccurate_ablation}, inaccurate pose features caused larger errors than removing these features. However, torque features worked conversely. Thus, a further experiment focusing on motor torque and current command was performed and the result is shown in Table \ref{tab_torque_ablation}. In this experiment, torque features were modified by 4 different operations, and the influence on calibration error was studied. In Table \ref{tab_torque_ablation}, 'None' means the original torque features were used with no modification, '$\pm$ STDV' means that white noise with range $\pm$ STDV was added. These two trials were used as baseline comparisons. The rest two trials, '$\times$ -1' and '$\times$ 3' means multiplying the torque features with -1 and 3, which resulted in the same change in amplitude. The increase in error with '$\times$ -1', which reversed the direction, was much larger than that with '$\times$ 3', which did not reverse the direction, especially in joint 3. This observation showed that in the learning calibration of this paper, the direction of torque features was more important than the amplitude of these features.

\begin{table}[h]
\caption{\vspace{-0.3em} Calibration Error under Modified Torque Features}
\vspace{-2.5em}
\begin{center}
\begin{tabular}{crrrrrr}
\hline
\multicolumn{1}{c|}{\multirow{2}{*}{Operation}} & \multicolumn{2}{c|}{Joint 1}                                    & \multicolumn{2}{c|}{Joint 2}                                    & \multicolumn{2}{c}{Joint 3}                         \\
\multicolumn{1}{c|}{}                           & \multicolumn{1}{c}{RSME} & \multicolumn{1}{c|}{Peak}            & \multicolumn{1}{c}{RSME} & \multicolumn{1}{c|}{Peak}            & \multicolumn{1}{c}{RSME} & \multicolumn{1}{c}{Peak} \\ \hline
\multicolumn{7}{c}{No Load}                                                                                                                                                                                                               \\ \hline
\multicolumn{1}{c|}{None}                       & 0.3003                   & \multicolumn{1}{r|}{0.5902}          & 0.2888                   & \multicolumn{1}{r|}{1.2669}          & 0.1565                   & 0.6409                   \\
\multicolumn{1}{c|}{$\pm$ STDV}                    & 0.3163                   & \multicolumn{1}{r|}{0.6227}          & 0.3425                   & \multicolumn{1}{r|}{1.5971}          & 0.1635                   & 0.7677                   \\
\multicolumn{1}{c|}{\textbf{$\times$ -1}}              & \textbf{0.3881}          & \multicolumn{1}{r|}{\textbf{0.7614}} & \textbf{0.6034}          & \multicolumn{1}{r|}{\textbf{1.7910}} & \textbf{0.8395}          & \textbf{1.3886}          \\
\multicolumn{1}{c|}{$\times$ 3}                        & 0.2944                   & \multicolumn{1}{r|}{0.6077}          & 0.4873                   & \multicolumn{1}{r|}{1.4329}          & 0.2174                   & 0.6940                   \\ \hline
\multicolumn{7}{c}{500g Load}                                                                                                                                                                                                             \\ \hline
\multicolumn{1}{c|}{None}                       & 0.4456                   & \multicolumn{1}{r|}{0.7721}          & 0.3052                   & \multicolumn{1}{r|}{1.4299}          & 0.1900                   & 0.7513                   \\
\multicolumn{1}{c|}{$\pm$ STDV}                    & 0.4656                   & \multicolumn{1}{r|}{0.8259}          & 0.3901                   & \multicolumn{1}{r|}{1.7438}          & 0.1952                   & 0.8456                   \\
\multicolumn{1}{c|}{\textbf{$\times$ -1}}              & \textbf{0.5285}          & \multicolumn{1}{r|}{\textbf{0.9855}} & \textbf{0.5946}          & \multicolumn{1}{r|}{\textbf{1.7524}} & \textbf{0.7608}          & \textbf{1.4761}          \\
\multicolumn{1}{c|}{$\times$ 3}                        & 0.4477                   & \multicolumn{1}{r|}{0.8249}          & 0.5018                   & \multicolumn{1}{r|}{1.7231}          & 0.2186                   & 0.6975                   \\ \hline
\end{tabular}

\end{center}
\vspace{-3.0em}
\label{tab_torque_ablation}
\end{table}

\subsection{Homing Inconsistency} \label{schar_homing_inconsist}
As mentioned in \ref{schar_exp_setup}, homing is an important procedure during robot initialization after power on, in which the robot explores the hard limit of each joint and register the encoders. To study the influences of robot homing, a special data set was collected. The training set was collected in the same way as datasets in \ref{sch_datasets}. The test set, while also consisting of several random sinusoidal trajectories, re-homing was applied to the robot in between the test trajectories (hardware power was not restarted). The first test trajectory was collected right after the training trajectories with no homing.

\begin{figure}
\centering
\vspace{0.5em}
\includegraphics[width=0.45\textwidth]{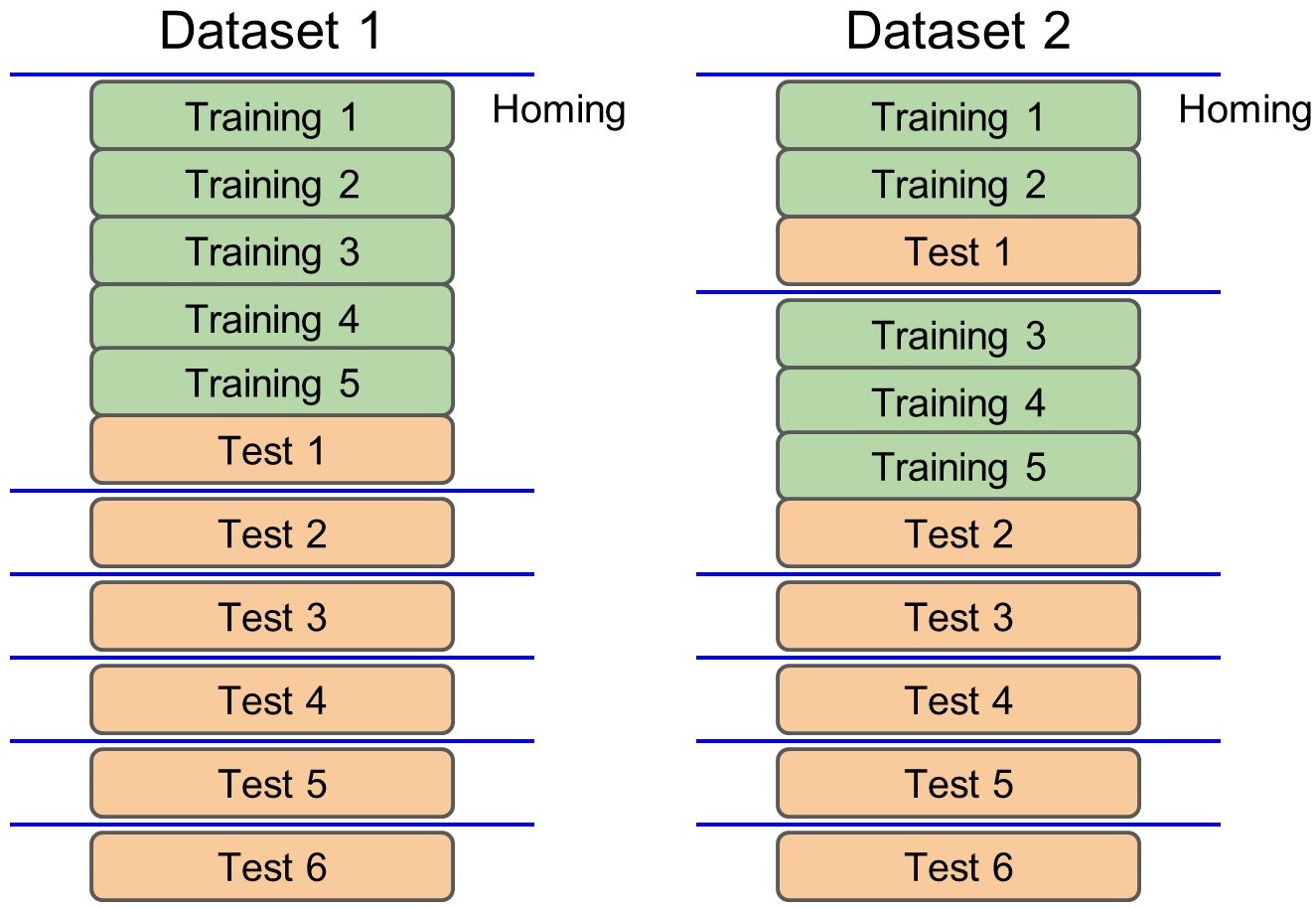}
\vspace{-1.em}
\caption{Two datasets, without homing in training (left), and with homing in training(right).} 
\vspace{-0.5em}
\label{data_w_homing}
\end{figure}

\begin{figure}
\centering
\vspace{0.5em}
\includegraphics[width=0.48\textwidth]{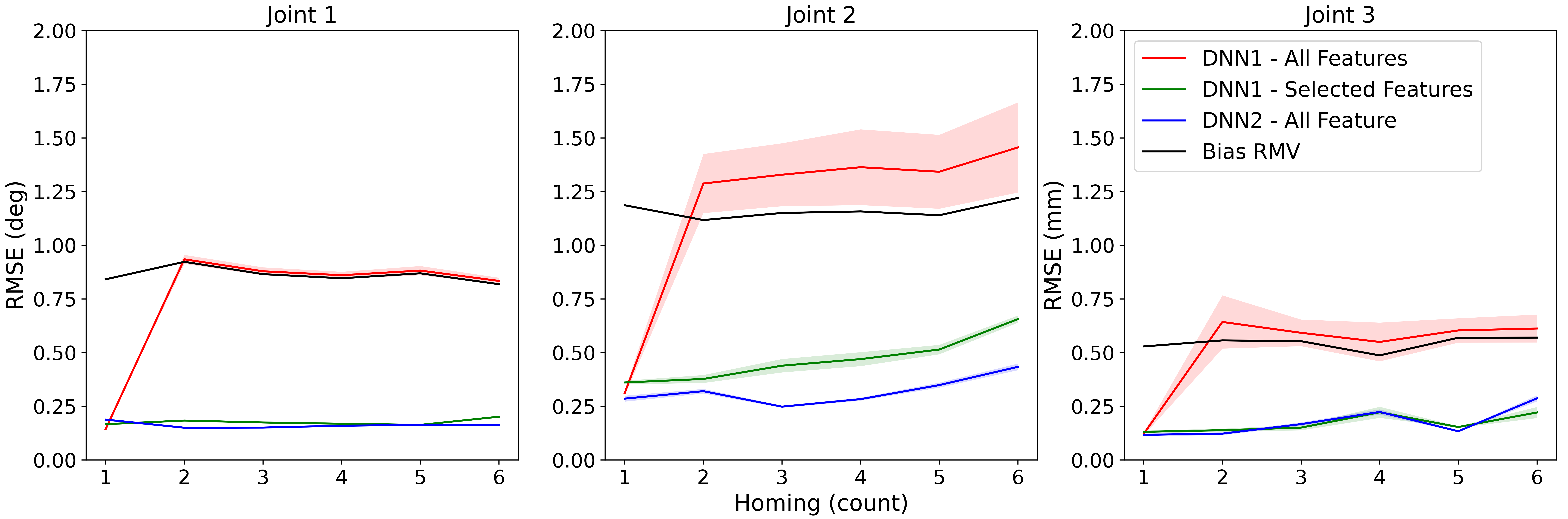}
\vspace{-1.em}
\caption{Robustness of calibration error against robot homing when using different features as input of the DNN model. 6 individual test sets were collected with robot homing in between. The first test set was collected right after the training set without homing. In selected features, encoder values \& offsets and a void motor pose were excluded. To better compare the improvement of learning calibration, error after a bias compensation was used as baseline (black), instead of the error of the original joint poses in robot states.} 
\vspace{-0.5em}
\label{homing_decay}
\end{figure}

\begin{figure}
\centering
\vspace{0.5em}
\includegraphics[width=0.48\textwidth]{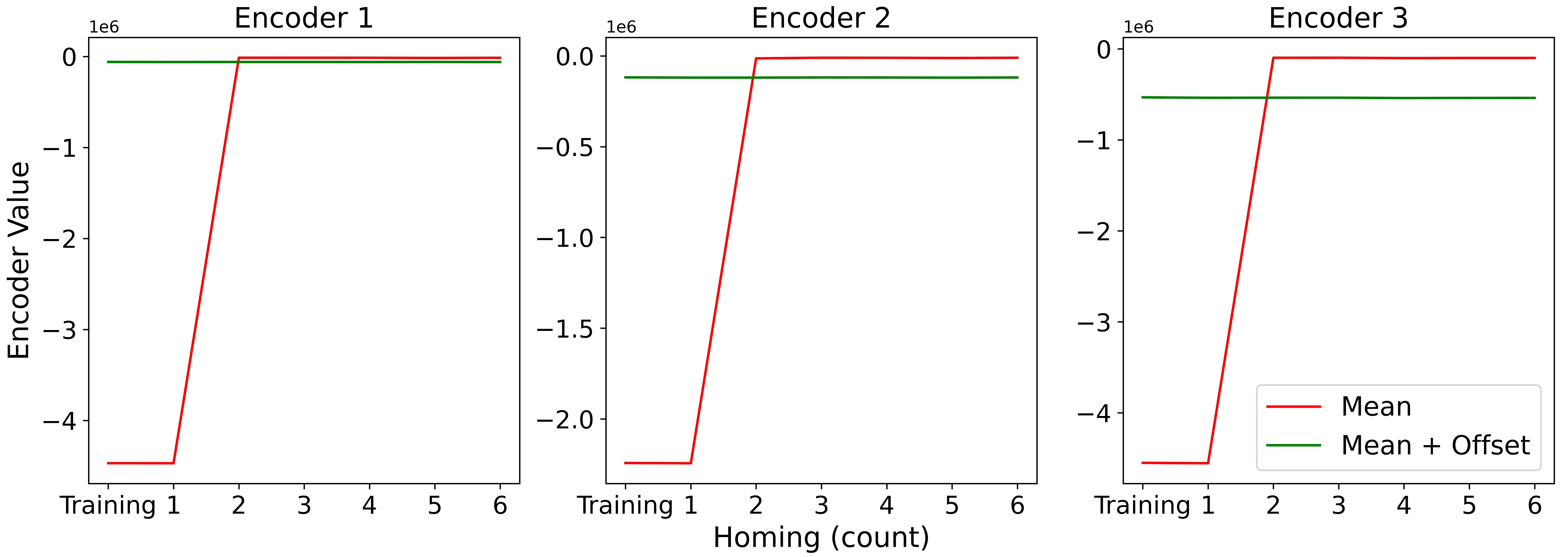}
\vspace{-1.em}
\caption{Encoder values affected by robot homing. The mean encoder value and offset compensated mean encoder value of the training set and 6 test sets after 0-5 robot homing are shown. The power of robot hardware stays on during the collection of training and test sets.} 
\vspace{-0.5em}
\label{encoder_value}
\end{figure}

As Fig. \ref{homing_decay} shows, if all the features in the robot state were used, the calibration errors were small in all 3 joints before the robot homing. However, after at least 1 homing, the error increase dramatically. And the error did not increase considerably as the homing time increased. Removal ablation was used to find which features caused the increase in calibration error. After attempts, 3 features were found causing the large error. The first 2 features were encoder values and encoder offsets, shown in Fig. \ref{encoder_value}. It can be seen that after the first homing, the average encoder values in the trajectories changed significantly, but further homings did not change these values considerably. However, motor poses and joint poses stayed consistent regardless of homings. This was because when adding encoder offset back to encoder values, these values also stayed consistent. But the DNN model failed to learn this linear combination of features.

The last feature was the 'motor pose 4' in ROS topic 'ravenstate'. This scatter number does not represent the actual pose of the 4th motor. Instead, this is a meaningless feature in our experiments, since the RAVEN-II has a spare independent joint dimension and is not used with the standard RAVEN tool. In the homing experiment, this feature followed a similar pattern to encoder values and had a rapid change after the first homing.

\begin{table*}[h]
\caption{\vspace{-0.3em} Calibration Error under modified pose Features}
\vspace{-1.5em}
\begin{center}
\begin{tabular}{crrrrrr}
\hline
\multicolumn{1}{c|}{\multirow{2}{*}{Operation}}                                                    & \multicolumn{2}{c|}{Joint 1}                           & \multicolumn{2}{c|}{Joint 2}                           & \multicolumn{2}{c}{Joint 3}                         \\
\multicolumn{1}{c|}{}                                                                              & \multicolumn{1}{c}{RSME} & \multicolumn{1}{c|}{Peak}   & \multicolumn{1}{c}{RSME} & \multicolumn{1}{c|}{Peak}   & \multicolumn{1}{c}{RSME} & \multicolumn{1}{c}{Peak} \\ \hline
\multicolumn{7}{c}{Dataset 1 - No homing in training data}                                                                                                                                                                                                                 \\ \hline
\multicolumn{1}{c|}{All features}                                                                  & 0.7554                   & \multicolumn{1}{r|}{1.6125} & 1.1813                   & \multicolumn{1}{r|}{3.5517} & 0.5201                   & 1.4590                   \\
\multicolumn{1}{c|}{No encoder}                                                                    & 0.1762                   & \multicolumn{1}{r|}{0.5628} & 0.4694                   & \multicolumn{1}{r|}{1.7209} & 0.1693                   & 0.6448                   \\ \hline
\multicolumn{7}{c}{Dataset 2 - Homing in training data}                                                                                                                                                                                                                    \\ \hline
\multicolumn{1}{c|}{All features}                                                                  & 0.1620                   & \multicolumn{1}{r|}{0.5526} & 0.3198                   & \multicolumn{1}{r|}{1.2084} & 0.1750                   & 0.6181                   \\
\multicolumn{1}{c|}{Inaccurate Encoder}                                                            & 0.1532                   & \multicolumn{1}{r|}{0.5029} & 0.3193                   & \multicolumn{1}{r|}{1.2600} & 0.1748                   & 0.5956                   \\
\multicolumn{1}{c|}{\begin{tabular}[c]{@{}c@{}}No other poses, \\ only encoder\end{tabular}}       & 0.1477                   & \multicolumn{1}{r|}{0.4372} & 0.2907                   & \multicolumn{1}{r|}{1.3735} & 0.1722                   & 0.6448                   \\
\multicolumn{1}{c|}{\begin{tabular}[c]{@{}c@{}}No other poses, \\ inaccurate encoder\end{tabular}} & 0.4343                   & \multicolumn{1}{r|}{1.2983} & 0.6191                   & \multicolumn{1}{r|}{2.4136} & 0.2106                   & 0.8770                   \\ \hline
\end{tabular}
\end{center}
\vspace{-3.0em}
\label{tab_homing_ablation}
\end{table*}

\section{Discussion}

Although ablation studies were performed, the DNN model itself remains a 'black box' and it may not be correct to speculate how DNN accomplished the tasks using human understanding on the tasks. Thus, the discussion in this section is rather suggested than conclusive.

From Table \ref{tab_removal_ablation} and \ref{tab_inaccurate_ablation}, it can be seen that the removal of either joint pose features or end-effector features did not undermine the performance of the learning calibration significantly. A small improvement could even be found in joint 3. However, once both joint and end-effector pose features were removed, significant increases in errors of all 3 joints were observed, especially in joints 1 and 2. On the other hand, once joint pose features were inaccurate, errors increased significantly, which suggested that joint pose features were important for the learning calibration of joints. Recall that removal ablation removed features in both training and testing, while inaccurate ablation only added noise when testing. Thus, the comparison of these ablation trials indicated that the DNN model relied more on joint pose information than end-effector information, for joint calibration. However if joint pose features were not provided during training, the DNN model was 'forced' or 'guided' to derive joint features based on end-effector features. This may reveal the potential that this DNN model, though not very deep with many hidden units, could solve inverse kinematics, which is highly non-linear for a 7-joint robot arm, without knowing DH parameters.

We already find that joint pose features were important, but DNN can also derive the joint information using end-effector poses. However, when both types of features were provided, inaccurate joint pose features caused much more error than inaccurate end-effector features. It may be suggested that joint pose features had a more direct relationship to joint calibration, and were thus given a larger weight during training. This may reveal that the DNN model was 'lazy', once there were direct and easy-to-use features, the model did not tend to learn indirect features.

From the ablation study, inaccuracy pose features caused larger errors than missing pose features. This suggested that in learning calibration practice if certain features are expected to be inaccurate, it may be a good idea to exclude this feature in training to 'force' or 'guide' the learning model to derive the information of these features from other features. However, the supplementary features should have explicit or implicit relations with the removed features, which can be highly non-linear.

It can also be seen from Table \ref{tab_removal_ablation} and \ref{tab_inaccurate_ablation} that the removal of joint 1 features did not drop the accuracy of joint 1 but inaccurate joint 1 features did, which indicated that joint 1 information was critical to the learning calibration of joint 1. But the DNN model could also derive this information from the other 2 joints. On the other hand, missing joint 2 features undermined the performance of both joints 2 and 3, but inaccurate joint 2 features only undermined the performance of joint 2. This suggested that joint 2 information was important to the calibration of joint 2 and 3, while joint 3 calibration was robust on the inaccuracy of joint 2 information. In contrast, either removal or inaccuracy of joint 3 features increased error on any joints, which indicated that joint 3 information was either not relative to the learning calibration, or the DNN model had robustness on inaccurate information of joint 3.

\section{Conclusions and Future Work}
In this paper, a learning-based calibration of RAVEN-II surgical robot is developed, utilizing a DNN model that takes the robot's original state estimation as input and output calibrated joint poses. After trained with zig-zag trajectories, the DNN reduced at least 85\% (unloaded) and 79\% (loaded) of joint errors. To study the contribution of the input features, removal ablation and inaccurate ablation are performed. The result of the 2 ablation studies shows that raw joint poses before calibration, as well as motor torques, are the most critical features to ensure calibration accuracy. Inaccurate raw joint poses undermine the accuracy dramatically. However, it is possible to guide the DNN model to learn to derive joint poses from end-effector pose and orientation by removing joint poses in training. By modifying torque features, it is found that the direction of torques is more important than the amplitude. Inconsistency of encoder values after robot homing is also observed to cause a significant increase in calibration error, though encoder offset is also included in the features. The solution is to exclude these features without compromised performance, if joint pose features are provided. In terms of joint coupling, removal of joint 2 features cause increased error in both joint 2 and 3, while inaccurate joint 1 or 2 features only have influence on the accuracy of themselves.

In the future, it is interesting to further explore how to guide the machine learning model to utilize redundant features and improve robustness to noise.
We also plan to get inspiration on how learning-based methods use the features, to guide and refine model-based methods.

\bibliographystyle{IEEEtran}
\bibliography{IEEEabrv,IEEEexample}

\begin{thebibliography}{10}
\providecommand{\url}[1]{#1}
\csname url@rmstyle\endcsname
\providecommand{\newblock}{\relax}
\providecommand{\bibinfo}[2]{#2}
\providecommand\BIBentrySTDinterwordspacing{\spaceskip=0pt\relax}
\providecommand\BIBentryALTinterwordstretchfactor{4}
\providecommand\BIBentryALTinterwordspacing{\spaceskip=\fontdimen2\font plus
\BIBentryALTinterwordstretchfactor\fontdimen3\font minus
  \fontdimen4\font\relax}
\providecommand\BIBforeignlanguage[2]{{%
\expandafter\ifx\csname l@#1\endcsname\relax
\typeout{** WARNING: IEEEtran.bst: No hyphenation pattern has been}%
\typeout{** loaded for the language `#1'. Using the pattern for}%
\typeout{** the default language instead.}%
\else
\language=\csname l@#1\endcsname
\fi
#2}}

\bibitem{palep2009robotic}
J.~H. Palep, ``Robotic assisted minimally invasive surgery,'' \emph{Journal of
  minimal access surgery}, vol.~5, no.~1, p.~1, 2009.

\bibitem{sayari2019review}
A.~J. Sayari, C.~Pardo, B.~A. Basques, \emph{et~al.}, ``Review of
  robotic-assisted surgery: what the future looks like through a spine oncology
  lens,'' \emph{Annals of translational medicine}, vol.~7, no.~10, 2019.

\bibitem{peters2018review}
B.~S. Peters, P.~R. Armijo, C.~Krause, \emph{et~al.}, ``Review of emerging
  surgical robotic technology,'' \emph{Surgical endoscopy}, vol.~32, no.~4, pp.
  1636--1655, 2018.

\bibitem{camarillo2004robotic}
D.~B. Camarillo, T.~M. Krummel, and J.~K. Salisbury~Jr, ``Robotic technology in
  surgery: past, present, and future,'' \emph{The American Journal of Surgery},
  vol. 188, no.~4, pp. 2--15, 2004.

\bibitem{hannaford2012raven}
B.~Hannaford, J.~Rosen, D.~W. Friedman, \emph{et~al.}, ``Raven-ii: an open
  platform for surgical robotics research,'' \emph{IEEE Transactions on
  Biomedical Engineering}, vol.~60, no.~4, pp. 954--959, 2012.

\bibitem{kazanzides2014open}
P.~Kazanzides, Z.~Chen, A.~Deguet, \emph{et~al.}, ``An open-source research kit
  for the da vinci{\textregistered} surgical system,'' in \emph{2014 IEEE
  international conference on robotics and automation (ICRA)}, pp.
  6434--6439.\hskip 1em plus 0.5em minus 0.4em\relax IEEE, 2014.

\bibitem{seita2018fast}
D.~Seita, S.~Krishnan, R.~Fox, \emph{et~al.}, ``Fast and reliable autonomous
  surgical debridement with cable-driven robots using a two-phase calibration
  procedure,'' in \emph{2018 IEEE International Conference on Robotics and
  Automation (ICRA)}, pp. 6651--6658.\hskip 1em plus 0.5em minus 0.4em\relax
  IEEE, 2018.

\bibitem{peng2020real}
H.~Peng, X.~Yang, Y.-H. Su, \emph{et~al.}, ``Real-time data driven precision
  estimator for raven-ii surgical robot end effector position,'' in \emph{2020
  IEEE International Conference on Robotics and Automation (ICRA)}, pp.
  350--356.\hskip 1em plus 0.5em minus 0.4em\relax IEEE, 2020.

\bibitem{haghighipanah2015improving}
M.~Haghighipanah, Y.~Li, M.~Miyasaka, \emph{et~al.}, ``Improving position
  precision of a servo-controlled elastic cable driven surgical robot using
  unscented kalman filter,'' in \emph{2015 IEEE/RSJ international conference on
  intelligent robots and systems (IROS)}, pp. 2030--2036.\hskip 1em plus 0.5em
  minus 0.4em\relax IEEE, 2015.

\bibitem{miyasaka2016hysteresis}
M.~Miyasaka, M.~Haghighipanah, Y.~Li, \emph{et~al.}, ``Hysteresis model of
  longitudinally loaded cable for cable driven robots and identification of the
  parameters,'' in \emph{2016 IEEE International Conference on Robotics and
  Automation (ICRA)}, pp. 4051--4057.\hskip 1em plus 0.5em minus 0.4em\relax
  IEEE, 2016.

\bibitem{hwang2020efficiently}
M.~Hwang, B.~Thananjeyan, S.~Paradis, \emph{et~al.}, ``Efficiently calibrating
  cable-driven surgical robots with rgbd fiducial sensing and recurrent neural
  networks,'' \emph{IEEE Robotics and Automation Letters}, vol.~5, no.~4, pp.
  5937--5944, 2020.

\bibitem{hwang2022automating}
M.~Hwang, J.~Ichnowski, B.~Thananjeyan, \emph{et~al.}, ``Automating surgical
  peg transfer: Calibration with deep learning can exceed speed, accuracy, and
  consistency of humans,'' \emph{IEEE Transactions on Automation Science and
  Engineering}, 2022.

\bibitem{mahler2014learning}
J.~Mahler, S.~Krishnan, M.~Laskey, \emph{et~al.}, ``Learning accurate kinematic
  control of cable-driven surgical robots using data cleaning and gaussian
  process regression,'' in \emph{2014 IEEE international conference on
  automation science and engineering (CASE)}, pp. 532--539.\hskip 1em plus
  0.5em minus 0.4em\relax IEEE, 2014.

\bibitem{franzini2019ablative}
A.~Franzini, S.~Moosa, D.~Servello, \emph{et~al.}, ``Ablative brain surgery: an
  overview,'' \emph{International Journal of Hyperthermia}, vol.~36, no.~2, pp.
  64--80, 2019.

\bibitem{cohen1989toward}
P.~R. Cohen and A.~E. Howe, ``Toward ai research methodology: Three case
  studies in evaluation,'' \emph{IEEE Transactions on Systems, Man, and
  Cybernetics}, vol.~19, no.~3, pp. 634--646, 1989.

\bibitem{newell1975tutorial}
A.~Newell, ``A tutorial on speech understanding systems,'' \emph{Speech
  recognition}, pp. 4--54, 1975.

\bibitem{meyes2019ablation}
R.~Meyes, M.~Lu, C.~W. de~Puiseau, \emph{et~al.}, ``Ablation studies in
  artificial neural networks,'' \emph{arXiv preprint arXiv:1901.08644}, 2019.

\bibitem{kauchak2014text}
D.~Kauchak, O.~Mouradi, C.~Pentoney, \emph{et~al.}, ``Text simplification
  tools: Using machine learning to discover features that identify difficult
  text,'' in \emph{2014 47th Hawaii international conference on system
  sciences}, pp. 2616--2625.\hskip 1em plus 0.5em minus 0.4em\relax IEEE, 2014.

\bibitem{kingma2014adam}
D.~P. Kingma and J.~Ba, ``Adam: A method for stochastic optimization,''
  \emph{arXiv preprint arXiv:1412.6980}, 2014.

\bibitem{reddi2019convergence}
S.~J. Reddi, S.~Kale, and S.~Kumar, ``On the convergence of adam and beyond,''
  \emph{arXiv preprint arXiv:1904.09237}, 2019.

\bibitem{su2020collaborative}
Y.-H. Su, A.~Munawar, A.~Deguet, \emph{et~al.}, ``Collaborative robotics
  toolkit (crtk): Open software framework for surgical robotics research,'' in
  \emph{2020 Fourth IEEE International Conference on Robotic Computing (IRC)},
  pp. 48--55.\hskip 1em plus 0.5em minus 0.4em\relax IEEE, 2020.

\end{thebibliography}


\end{document}